\documentclass{article}
\usepackage{spconf,amsmath,graphicx}
\usepackage{verbatim}

\usepackage{url}
\usepackage{esvect}
\usepackage[usenames,dvipsnames,table]{xcolor}
\usepackage{amsmath}
\usepackage{amssymb}
\usepackage{mathtools}
\usepackage{hyperref}

\definecolor{darkblue}{RGB}{0, 76, 153}
\hypersetup{
    colorlinks,
    linkcolor={blue!50!black},
    citecolor={blue!50!black},
    urlcolor={blue!50!black}
}
\usepackage{algpseudocode}
\usepackage{pdflscape}
\usepackage{titlesec}

\usepackage{algorithm}
\usepackage{setspace}
\usepackage{booktabs}
\usepackage{subcaption}
\usepackage{tabularray}
\usepackage{tikz}
\usepackage{multirow}
\usepackage{booktabs}
\usepackage[capitalise]{cleveref}
\usetikzlibrary{bayesnet, calc, arrows.meta, bending}
\usetikzlibrary{decorations.pathreplacing,angles,quotes,shapes.multipart, calligraphy}

\newcommand{\unint}[1]{\textcolor{Bittersweet}{#1}}
\newcommand{\inter}[1]{\textcolor{RoyalBlue}{#1}}

\usepackage{multirow}

\usepackage{enumitem}
\setlist{nosep, leftmargin=14pt}

\usepackage{mwe} 

\title{Positional Segmentor-Guided Counterfactual Fine-Tuning for Spatially Localized Image Synthesis}

\name{%
\begin{tabular}{c}
Tian Xia$^{1}$, Matthew Sinclair$^{1,2}$, Andreas Schuh$^{1,2}$, Fabio De Sousa Ribeiro$^{1}$, Raghav Mehta$^{1}$, Rajat Rasal$^{1}$,\\
Esther Puyol-Ant\'{o}n$^{1,2}$, Samuel Gerber$^{2}$, Kersten Petersen$^{2}$, Michiel Schaap$^{1,2}$, and Ben Glocker$^{1}$\thanks{Corresponding author: t.xia@imperial.ac.uk}
\end{tabular}
}

\address{%
$^{1}$ Department of Computing, Imperial College London, UK \\
$^{2}$ HeartFlow, Inc., Mountain View, CA, USA
}

\begin{document}

\maketitle
\begin{abstract}
Counterfactual image generation enables controlled data augmentation, bias mitigation, and disease modeling. However, existing methods guided by external classifiers or regressors are limited to subject-level factors (e.g., age) and fail to produce localized structural changes, often resulting in global artifacts. Pixel-level guidance using segmentation masks has been explored, but requires user-defined counterfactual masks, which are tedious and impractical. Segmentor-guided Counterfactual Fine-Tuning (Seg-CFT) addressed this by using segmentation-derived measurements to supervise structure-specific variables, yet it remains restricted to global interventions. We propose Positional Seg-CFT, which subdivides each structure into regional segments and derives independent measurements per region, enabling spatially localized and anatomically coherent counterfactuals. Experiments on coronary CT angiography show that Pos-Seg-CFT generates realistic, region-specific modifications, providing finer spatial control for modeling disease progression.
\end{abstract}
\section{Introduction}
\label{sec:intro}

Causal questions such as ``\textit{How would this patient’s disease have progressed if treatment A had been administered instead of treatment B}?'' are central to scientific inquiry and clinical decision-making. Addressing such questions requires models that can simulate realistic interventions beyond statistical correlations. Counterfactual image generation enables this capability in medical imaging, supporting applications in data augmentation~\cite{ilse2021selecting,roschewitz2024robust}, bias mitigation~\cite{kumar2023debiasing}, explainability~\cite{pegios2024diffusion}, and disease progression modeling~\cite{puglisi2024enhancing}.


Previous works~\cite{yang2021causalvae,sanchez2022diffusion,geffner2022deep} tried to integrate causality with deep generative models~\cite{goodfellow2020generative,kingma2013auto,ho2020denoising}. However, most of them focused on association or intervention, without a principled approach to counterfactual reasoning, i.e., the highest level in Pearl's causal hierarchy. Notable exceptions include Neural Causal Models (NCMs) \cite{xia2021causal,xia2023neural,pan2024counterfactual} and Deep Structural Causal Models (DSCMs) \cite{pawlowski2020deep,de2023high,xia2024mitigating,rasal2025diffusion}, which integrate causal structures with deep generative models.

Ribeiro et al.~\cite{de2023high} trained Deep Structural Causal Models (DSCMs) using hierarchical VAEs conditioned on causal parents, but standard likelihood training often ignored interventions, violating counterfactual consistency~\cite{monteiro2023measuring,xia2024mitigating}. Counterfactual Fine-Tuning (CFT)~\cite{de2023high} addressed this by refining DSCMs with pretrained regressors to improve causal adherence. However, most previous works~\cite{de2023high,xia2024mitigating,roschewitz2024robust,roschewitz2024counterfactual,ibrahim2024semi,mehta2025cf} focused on patient-level factors (e.g., sex, age), while structure-specific interventions~\cite{xia2025segmentor} revealed that regression-based CFT (Reg-CFT) produces global rather than localized changes. This aligns with evidence that medical image regressors rely on spurious or non-local cues, motivating the use of segmentation-based spatial supervision.

While Seg-CFT improves local coherence, it relies on global measurements (e.g., total plaque area) and cannot specify where within the image changes occur. One alternative is to use segmentation masks to guide generative models~\cite{alaya2024mededit}. However, incorporating masks into a causal framework remains challenging: (i) their causal role is ambiguous, and (ii) reliance on predefined counterfactual masks limits practicality, as such masks are tedious to obtain.

In this work, we propose \textit{Positional Segmentor-guided Counterfactual Fine-Tuning (Pos-Seg-CFT)}, a simple yet effective extension of Seg-CFT that replaces global measurements with region-specific variables capturing structure changes within defined image segments (e.g., plaque area in proximal, mid, and distal regions). Pos-Seg-CFT enables spatially targeted interventions while preserving simple scalar supervision and avoiding pixel-level inputs. Using pretrained, weight-frozen segmentors, regional measurements are derived by masking other regions, allowing the same model to be reused. The approach is flexible and applicable to arbitrary region definitions, achieving fine-grained spatial control without predefined counterfactual masks. We validate Pos-Seg-CFT on coronary CT angiography (CCTA), demonstrating realistic, region-specific counterfactuals that capture localized patterns of disease progression.

\section{Background}

\paragraph{Structural Causal Models (SCMs)}  comprise a triplet $\langle U, V, F \rangle$~\cite{pearl2009causality}, where $U {=} \{u_{i}\}_{i=1}^K$ are a set of exogenous variables, $V{=}\{v_{i}\}_{i=1}^K$ a set of endogenous variables, and $F {=} \{f_{i}\}_{i=1}^K$ a set of functions such that  $v_k {\coloneqq} f_k(\mathbf{pa}_k, u_k)$, where  $\mathbf{pa}_{k}\subseteq V\setminus v_{k}$ are called direct causes or parents of $v_{k}$. Interventions are represented by the do-operator, which enforces modifications to one or more parent variables. Counterfactual inference follows three steps: (i) Abduction: estimating exogenous noise variables from observed data; (ii) Action: applying an intervention, e.g., $do(v_k {\coloneqq} c)$; and (iii) Prediction: generating counterfactual outcomes using the modified model and inferred exogenous variables.

\paragraph{Deep Structural Causal Models (DSCMs)} were first proposed in \cite{pawlowski2020deep} and later extended in \cite{de2023high} for high-resolution counterfactual image generation. Given an image $\mathbf{x}$, let $\{v_{1}, \dots, v_{K-1}\} \supseteq \mathbf{pa}_{\mathbf{x}}$ denote its \textit{ancestors}. Each low-dimensional attribute is described by an invertible conditional normalizing flow, $v_{k} = f_{k}(u_{k}; \mathbf{pa}_{k})$, making abduction explicit and tractable. For high-dimensional variables such as images, the generative mechanism is realized with an HVAE. To obtain a counterfactual image, we first infer the exogenous noise of the image, $\mathbf{z} {\sim} q_{\phi}(\mathbf{z} \mid \mathbf{x}, \mathbf{pa}_{\mathbf{x}})$, where $q_{\phi}$ is the HVAE encoder, and likewise infer attribute noise via $u_{k}{=}f_k^{-1}(v_{k}; \mathbf{pa}_{k})$. An intervention is then performed by  $do(v_{i} {\coloneqq} c)$, i.e., sets $v_{i}$ to some target $c$. We can perform interventions on multiple attributes simultaneously. Using the abducted noise $u_{k}$, we compute counterfactual parent values $\widetilde{\mathbf{pa}}_{\mathbf{x}}$ and generate the counterfactual image as $\widetilde{\mathbf{x}}{=}g_{\theta}(\mathbf{z}, \widetilde{\mathbf{pa}}_{\mathbf{x}})$.


\paragraph{Reg-CFT.} Likelihood-based HVAE training alone has been found to cause ignored counterfactual conditioning~\cite{de2023high,xia2024mitigating}, where the generated image $\widetilde{\mathbf{x}}$ fails to reflect the intervened parent variables $\mathbf{\widetilde{pa}}_{\mathbf{x}}$. To address this, \textit{counterfactual fine-tuning (CFT)} was introduced~\cite{de2023high}. The central idea is to employ pre-trained classifiers or regressors $q_{\xi}(\mathbf{pa}_{\mathbf{x}} {\mid} \mathbf{x})$ and fine-tune the HVAE parameters ${\theta, \phi}$ by maximizing $\log q_{\xi}(\mathbf{\widetilde{pa}}_{\mathbf{x}} {\mid} \widetilde{\mathbf{x}})$, while keeping $\xi$ fixed. This additional fine-tuning step enforces that $\mathbf{\widetilde{pa}}_{\mathbf{x}}$ is predictable from $\widetilde{\mathbf{x}}$, encouraging DSCMs to generate counterfactual images that faithfully implement the intended interventions. Following \cite{xia2025segmentor}, we refer to CFT methods that use pretrained regressors or classifiers for fine-tuning as Reg-CFT. For simplicity, we use the term \textit{regressor} to denote both regressors and classifiers. 

In \cite{xia2025segmentor}, Reg-CFT was evaluated for structure-specific interventions using scalar variables derived from 2D medical images. The results showed that Reg-CFT often produced global rather than localized changes, suggesting that limited structural guidance makes it difficult for DSCMs to capture the true semantics of variables such as plaque area. The regressors tended to exploit spurious correlations (e.g., vessel brightness) instead of the intended anatomical cues, underscoring the need for spatially informed supervision during counterfactual fine-tuning.

\paragraph{Seg-CFT.} To this end, Xia et al.~\cite{xia2025segmentor} proposed Segmentor-guided Counterfactual Fine-Tuning that leverages a pretrained segmentation model during counterfactual fine-tuning. They retain all variables of interest as scalar-valued, similar to Reg-CFT, including both subject-level and structure-specific variables. The main idea is to predict segmentation maps from counterfactual images and compute scalar measurements (e.g., areas) by summing pixel-level label probabilities. These segmentation-derived measurements guide the fine-tuning process, ensuring that counterfactual generations remain spatially consistent with the intended interventions. However, Seg-CFT relies on global measurements (e.g., total plaque area) and cannot localize structural changes, whereas Pos-Seg-CFT introduces region-specific supervision for spatially targeted interventions.

\section{Positional Segmentor-guided Counterfactual Fine-Tuning}

In this paper, we propose a simple yet effective extension of Seg-CFT, termed Positional Segmentor-guided Counterfactual Fine-Tuning (Pos-Seg-CFT). The key idea is to further exploit the spatial information contained in the predicted segmentation maps. Rather than aggregating the total area of a structure (e.g., calcified plaque) across the entire image, Pos-Seg-CFT allows regional measurements to be computed within specific subregions of the image, thereby enabling explicit control over where the intervention should occur.

Formally, given a predicted segmentation map $\mathbf{\widehat{m}}$ produced by a pretrained segmentor $s_{\psi}$,
\begin{equation}
\mathbf{\widehat{m}} \sim s_{\psi}(\mathbf{\widehat{m}} \mid \widetilde{\mathbf{x}}),
\end{equation}
we define a set of spatial masks $\mathbf{R}_{1}, \mathbf{R}_{2}, \dots, \mathbf{R}_{K}$ that partition the image into $K$ disjoint regions (e.g., proximal, mid, and distal segments).
The regional measurement for the $k$-th region is then computed as:
\begin{equation}
\mathbf{\widehat{pa}}_{\mathbf{x}}^{(k)} = \sum_{(i,j) \in \mathbf{R}_k} \mathbf{\widehat{m}}_{i,j},
\end{equation}
which corresponds to the predicted area of the target structure within region $\mathbf{R}_k$.

These regional measurements are used as additional guidance signals during counterfactual fine-tuning, encouraging the generative model to produce spatially coherent modifications aligned with the intended positional interventions. Specifically, the fine-tuning objective minimizes the difference between the predicted and target regional measurements derived from the pretrained segmentor. 

\begin{equation}
\mathcal{L}_{\text{Pos-Seg-CFT}}(\theta, \phi)
= \sum_{k=1}^{K}
\mathcal{L}\big(\mathbf{\widehat{pa}}_{\mathbf{x}}^{(k)},\mathbf{\widetilde{pa}}_{\mathbf{x}}^{(k)}\big),
\end{equation}
\noindent where 
$\widehat{\mathbf{pa}}_{\mathbf{x}}^{(k)} {=}\sum_{(i,j)\in \mathbf{R}_k} \widehat{\mathbf{m}}_{i,j}$ 
denotes the predicted area of the target structure within region $\mathbf{R}_k$, 
$\widetilde{\mathbf{pa}}_{\mathbf{x}}^{(k)}$ is the target (counterfactual) regional measurement, 
and $\mathcal{L}(\cdot)$ is typically an $L_{1}$ or $L_{2}$ loss.
Regional measurements are obtained by reusing the same pretrained segmentor $s_{\psi}$, masking out other regions, and aggregating within the region of interest.
This design makes Pos-Seg-CFT a natural and flexible extension of Seg-CFT, enabling spatially localized counterfactuals without pixel-level supervision or extra annotations.
\begin{figure*}[t!]
    \centering
    \begin{subfigure}[t]{0.9\textwidth}
        \centering
        \includegraphics[width=\linewidth]{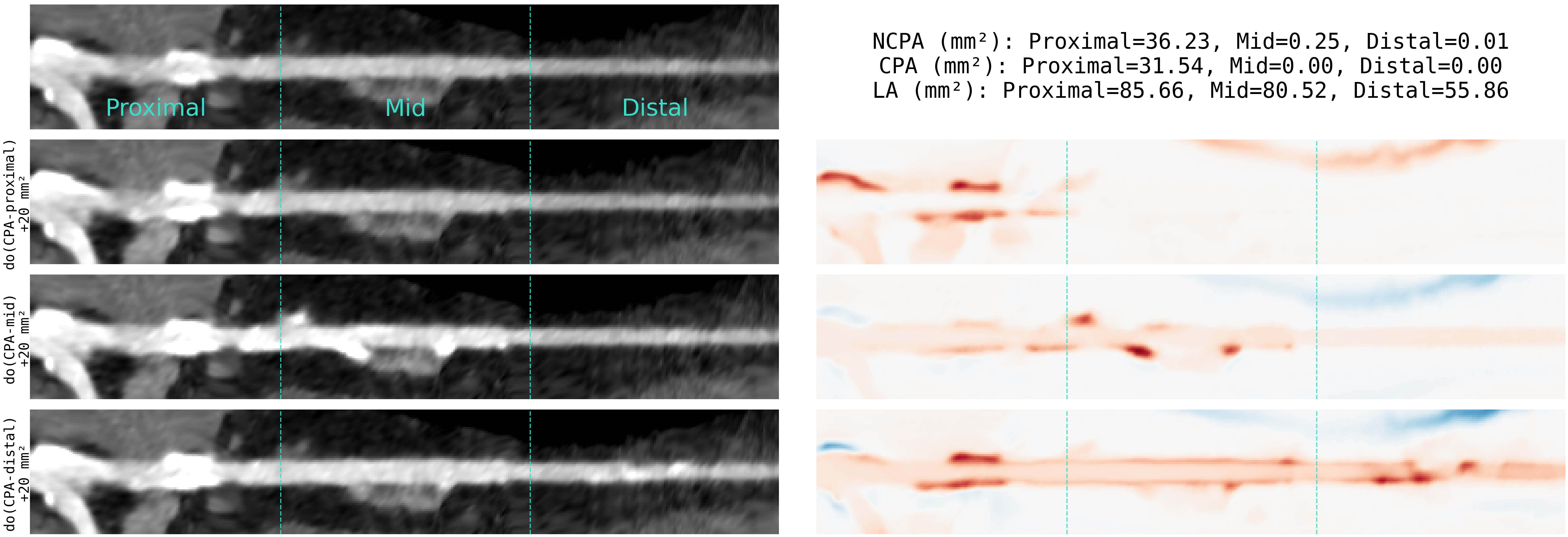}
        \caption{Reg-CFT.}
        \label{fig:Reg-CFT}
    \end{subfigure}
    \hspace{0.04\textwidth}
    \begin{subfigure}[t]{0.9\textwidth}
        \centering
        \includegraphics[width=\linewidth]{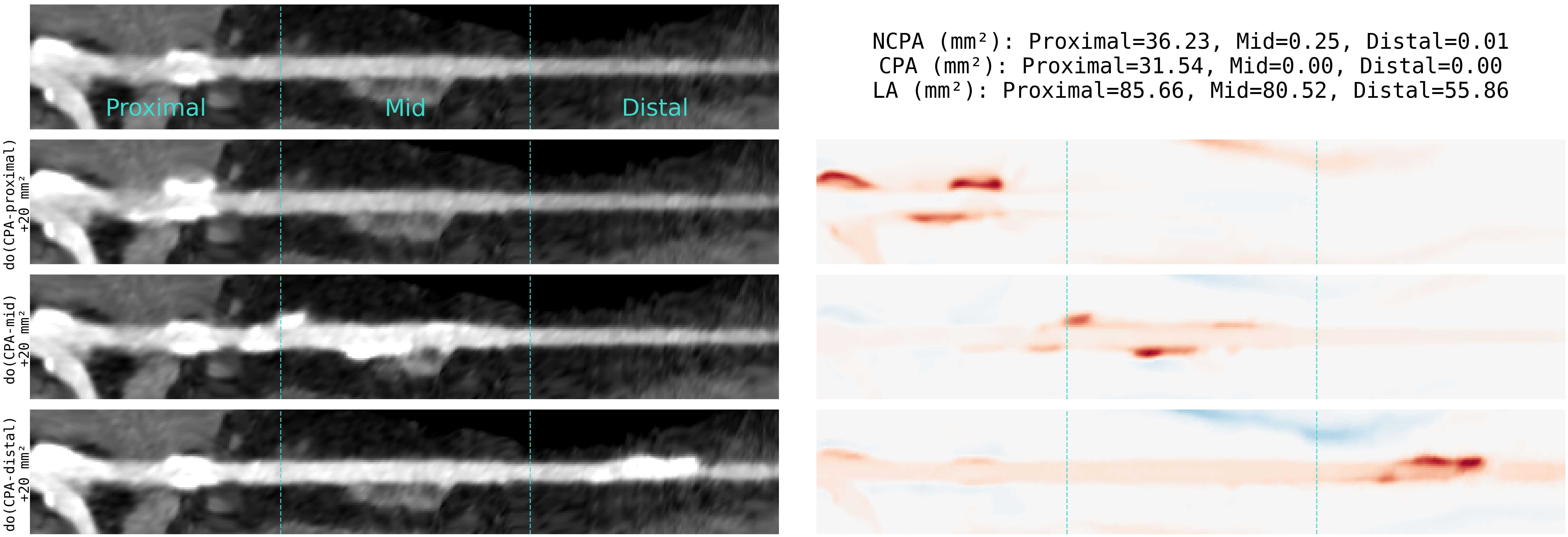}
        \caption{Pos-Seg-CFT.}
        \label{fig:pos-seg-cft}
    \end{subfigure}
\caption{Visual results under Reg-CFT and Pos-Seg-CFT for calcified plaque area (CPA) interventions of $+20$~mm$^2$ applied to the proximal, mid, and distal regions. Each row shows the counterfactual image (left) and the corresponding difference map (right). Vertical dashed lines indicate the \textit{proximal}, \textit{mid}, and \textit{distal} vessel segments.}
    \label{fig:pos-cft-comparison}
\end{figure*}

\section{Experiments \& Results}
\textbf{Experimental setup.} We conduct experiments on an internal coronary computed tomography angiography (CCTA) dataset~\cite{taylor2023}, an essential modality for evaluating coronary artery disease (CAD) through detailed visualization of plaque composition and distribution. Following the preprocessing protocol of~\cite{xia2025segmentor}, straightened curvilinear planar reformations (sCPR) of the left anterior descending (LAD) artery~\cite{kanitsar2002} were generated at a resolution of $0.25{\times}0.25$~mm and cropped to $64{\times}384$~pixels. Segmentation masks of the lumen, calcified plaque, and non-calcified plaque were obtained by rasterizing 3D meshes of the coronary lumen and outer wall onto the 2D sCPR plane. In total, 65,706 CCTA images were processed, with 49,292 for training, 6,565 for validation, and 9,849 for testing. For counterfactual modeling, we consider three structure-specific variables: calcified plaque area (CPA), non-calcified plaque area (NCPA), and lumen area (LA). To enable positional guidance in Pos-Seg-CFT, each image is further divided into proximal, mid, and distal regions along the vessel axis. Regional measurements are computed within each region, providing position-aware supervision for structure-specific interventions. For simplicity, all variables are treated as independent of each other. We compare the proposed \textit{Pos-Seg-CFT} with two baselines: \textit{No-CFT}~\cite{de2023high}, where no counterfactual fine-tuning is performed, and \textit{Reg-CFT}~\cite{de2023high}, which uses pre-trained ResNet-based regressors to predict the target regional measurements (e.g., CPA-proximal). Counterfactual performance is evaluated via \textit{effectiveness}~\cite{monteiro2023measuring}, which measures the alignment between the desired and predicted counterfactual parents, $d(\widehat{\mathbf{pa}}_{\mathbf{x}}, \widetilde{\mathbf{pa}}_{\mathbf{x}})$. The segmentor used for evaluation is trained independently from the one used for fine-tuning to ensure an unbiased assessment

\begin{table}[t!]
\centering
\scriptsize
\setlength{\tabcolsep}{4pt}
\caption{Quantitative \textit{effectiveness} of causal interventions $do(\text{variable–position})$ across positional regions (Proximal–Mid–Distal). Each block reports results for \textit{No-CFT}, \textit{Reg-CFT}, and the proposed \textit{Pos-Seg-CFT}. The \inter{blue} and \unint{orange} values denote the \inter{intervened} and \unint{unintervened} regions, respectively. \textbf{Bold} numbers indicate the best (lowest) error.}
\begin{tabular}{llccc}
\toprule
\textbf{Intervention} & \textbf{CFT type} & \textbf{Proximal} & \textbf{Mid} & \textbf{Distal} \\
\midrule
\multirow{3}{*}{do(NCPA-proximal)} 
 & No-CFT & \inter{29.50} & \unint{8.79} & \unint{3.25} \\
 & Reg-CFT & \inter{18.71} & \unint{8.01} & \unint{2.65} \\
 & Pos-Seg-CFT & \inter{\textbf{17.85}} & \unint{\textbf{7.27}} & \unint{\textbf{2.42}} \\
\midrule
\multirow{3}{*}{do(NCPA-mid)}
 & No-CFT & \unint{13.15} & \inter{37.22} & \unint{7.01} \\
 & Reg-CFT & \unint{12.82} & \inter{26.32} & \unint{6.15} \\
 & Pos-Seg-CFT & \unint{\textbf{11.50}} & \inter{\textbf{23.44}} & \unint{\textbf{4.91}} \\
\midrule
\multirow{3}{*}{do(NCPA-distal)}
 & No-CFT & \unint{12.48} & \unint{11.31} & \inter{43.36} \\
 & Reg-CFT & \unint{11.53} & \unint{10.87} & \inter{21.00} \\
 & Pos-Seg-CFT & \unint{\textbf{8.44}} & \unint{\textbf{7.34}} & \inter{\textbf{18.71}} \\
\midrule
\multirow{3}{*}{do(CPA-proximal)}
 & No-CFT & \inter{9.25} & \unint{3.70} & \unint{0.60} \\
 & Reg-CFT & \inter{7.89} & \unint{2.97} & \unint{0.60} \\
 & Pos-Seg-CFT & \inter{\textbf{7.77}} & \unint{\textbf{2.89}} & \unint{\textbf{0.55}} \\
\midrule
\multirow{3}{*}{do(CPA-mid)}
 & No-CFT & \unint{6.58} & \inter{14.64} & \unint{4.53} \\
 & Reg-CFT & \unint{5.03} & \inter{9.07} & \unint{2.90} \\
 & Pos-Seg-CFT & \unint{\textbf{4.88}} & \inter{\textbf{8.95}} & \unint{\textbf{2.41}} \\
\midrule
\multirow{3}{*}{do(CPA-distal)}
 & No-CFT & \unint{5.65} & \unint{13.69} & \inter{28.28} \\
 & Reg-CFT & \unint{4.36} & \unint{4.64} & \inter{12.14} \\
 & Pos-Seg-CFT & \unint{\textbf{3.97}} & \unint{\textbf{4.18}} & \inter{\textbf{9.85}} \\
\midrule
\multirow{3}{*}{do(LA-proximal)}
 & No-CFT & \inter{54.21} & \unint{14.45} & \unint{5.80} \\
 & Reg-CFT & \inter{28.91} & \unint{10.44} & \unint{\textbf{3.84}} \\
 & Pos-Seg-CFT & \inter{\textbf{24.25}} & \unint{\textbf{9.78}} & \unint{{3.93}} \\
\midrule
\multirow{3}{*}{do(LA-mid)}
 & No-CFT & \unint{32.61} & \inter{68.05} & \unint{26.20} \\
 & Reg-CFT & \unint{20.64} & \inter{39.92} & \unint{27.38} \\
 & Pos-Seg-CFT & \unint{\textbf{19.06}} & \inter{\textbf{22.52}} & \unint{\textbf{16.83}} \\
\midrule
\multirow{3}{*}{do(LA-distal)}
 & No-CFT & \unint{23.21} & \unint{29.56} & \inter{80.20} \\
 & Reg-CFT & \unint{11.32} & \unint{24.20} & \inter{66.76} \\
 & Pos-Seg-CFT & \unint{\textbf{8.86}} & \unint{\textbf{17.12}} & \inter{\textbf{19.38}} \\
\bottomrule
\end{tabular}
\label{tab: numeric results}
\end{table}

\textbf{Results.} Quantitative effectiveness is summarized in \cref{tab: numeric results}, reporting effect magnitudes across positional regions (Proximal–Mid–Distal) for NCPA, CPA, and LA. For each intervention, we randomly sample the counterfactual target value from the variable’s valid range (between its minimum and maximum observed values) to ensure consistent effect magnitudes across regions. Without fine-tuning (No-CFT), interventions on a target region (e.g., NCPA-proximal) often induce large off-target changes in adjacent or unrelated regions, reflecting poor localization and strong entanglement between anatomical components. The baseline Reg-CFT~\cite{de2023high}, which employs regression-based positional supervision, mitigates some of these off-target effects and yields more balanced magnitudes across positions. However, residual leakage remains—for instance, interventions on NCPA still slightly affect CPA and LA regions—indicating that regression alone does not fully disentangle spatial dependencies. In contrast, the proposed Pos-Seg-CFT achieves the most localized and stable effects, consistently reducing non-target activations while preserving strong targeted changes. This demonstrates that incorporating segmentation-guided positional supervision enhances causal precision and spatial specificity during counterfactual fine-tuning.

Visual results are shown in Fig.~\ref{fig:pos-cft-comparison}. The baseline Reg-CFT (Fig.~\ref{fig:pos-cft-comparison}a) exhibits noticeable positional leakage: interventions on \textit{CPA-mid} and \textit{CPA-distal} still trigger pronounced changes in the proximal segment, indicating that regression-based supervision fails to preserve spatial independence. In contrast, Pos-Seg-CFT (Fig.~\ref{fig:pos-cft-comparison}b) produces well-localized and anatomically consistent modifications confined to the intended regions. The difference maps further reveal spatially aligned responses and reduced off-target activations, demonstrating improved precision in positional counterfactual generation.

\section{Conclusion}
\label{sec:conclusion}

We presented {Positional Segmentor-guided Counterfactual Fine-Tuning (Pos-Seg-CFT)}, a novel extension of Seg-CFT that enables spatially localized and anatomically coherent counterfactual generation. By decomposing global structure-specific variables into regional measurements, Pos-Seg-CFT introduces position-aware supervision that provides fine-grained control over where causal interventions occur. During training, a pretrained, weight-frozen segmentor provides regional measurements to guide fine-tuning. Experiments on coronary CT angiography (CCTA) demonstrate that Pos-Seg-CFT substantially improves spatial localization, reduces off-target effects, and yields more stable and interpretable counterfactuals compared to regression-based fine-tuning. This framework offers a general approach for structure-aware causal modeling, paving the way toward anatomically grounded and interpretable generation in medical imaging.

\section{Compliance with ethical standards}
The CCTA dataset is based on secondary, fully anonymised data and exempt from ethical approval.

\section{Acknowledgments}
This research was funded by Heartflow, the Royal Academy of Engineering, EPSRC (grant no. EP/Y028856/1), and EU Horizon Europe programme (grant no.  101080302).

\bibliographystyle{IEEEtran}
\bibliography{refs}

@inproceedings{mehta2025cf,
  title={Cf-seg: Counterfactuals meet segmentation},
  author={Mehta, Raghav and De Sousa Ribeiro, Fabio and Xia, Tian and Roschewitz, M{\'e}lanie and Santhirasekaram, Ainkaran and Marshall, Dominic C and Glocker, Ben},
  booktitle={MICCAI},
  pages={117--127},
  year={2025},
  organization={Springer}
}

@article{de2023high,
  title={High Fidelity Image Counterfactuals with Probabilistic Causal Models},
  author={De Sousa Ribeiro, Fabio and Xia, Tian and Monteiro, Miguel and Pawlowski, Nick and Glocker, Ben},
  journal={ICML},
  year={2023}
}

@book{pearl2009causality,
  title={Causality},
  author={Pearl, Judea},
  year={2009},
  publisher={Cambridge university press}
}

@article{goodfellow2020generative,
  title={Generative adversarial networks},
  author={Goodfellow, Ian and Pouget-Abadie, Jean and Mirza, Mehdi and Xu, Bing and Warde-Farley, David and Ozair, Sherjil and Courville, Aaron and Bengio, Yoshua},
  journal={Communications of the ACM},
  volume={63},
  number={11},
  pages={139--144},
  year={2020},
  publisher={ACM New York, NY, USA}
}

@inproceedings{yang2021causalvae,
  title={CausalVAE: Disentangled representation learning via neural structural causal models},
  author={Yang, Mengyue and Liu, Furui and Chen, Zhitang and Shen, Xinwei and Hao, Jianye and Wang, Jun},
  booktitle={CVPR},
  pages={9593--9602},
  year={2021}
}

@article{geffner2022deep,
  title={Deep End-to-end Causal Inference},
  author={Geffner, Tomas and Antoran, Javier and Foster, Adam and Gong, Wenbo and Ma, Chao and Kiciman, Emre and Sharma, Amit and Lamb, Angus and Kukla, Martin and Pawlowski, Nick and others},
  journal={TMLR},
  year={2022}
}

@article{sanchez2022diffusion,
  title={Diffusion Models for Causal Discovery via Topological Ordering},
  author={Sanchez, Pedro and Liu, Xiao and O'Neil, Alison Q and Tsaftaris, Sotirios A},
  journal={ICLR},
  year={2023}
}

@article{xia2021causal,
  title={The causal-neural connection: Expressiveness, learnability, and inference},
  author={Xia, Kevin and Lee, Kai-Zhan and Bengio, Yoshua and Bareinboim, Elias},
  journal={NeurIPS},
  year={2021}
}

@inproceedings{
xia2023neural,
title={Neural Causal Models for Counterfactual Identification and Estimation},
author={Kevin Muyuan Xia and Yushu Pan and Elias Bareinboim},
booktitle={ ICLR },
year={2023},
}

@article{ho2020denoising,
  title={Denoising diffusion probabilistic models},
  author={Ho, Jonathan and Jain, Ajay and Abbeel, Pieter},
  journal={NeurIPS},
  year={2020}
}

@inproceedings{
monteiro2023measuring,
title={Measuring axiomatic soundness of counterfactual image models},
author={Miguel Monteiro and Fabio De Sousa Ribeiro and Nick Pawlowski and Daniel C. Castro and Ben Glocker},
booktitle={ICLR},
year={2023},
}

@article{pawlowski2020deep,
  title={Deep structural causal models for tractable counterfactual inference},
  author={Pawlowski, Nick and Coelho de Castro, Daniel and Glocker, Ben},
  journal={NeurIPs},
  volume={33},
  pages={857--869},
  year={2020}
}

@article{kingma2013auto,
  title={Auto-encoding variational bayes},
  author={Kingma, Diederik P and Welling, Max},
  journal={arXiv preprint arXiv:1312.6114},
  year={2013}
}

@inproceedings{alaya2024mededit,
  title={MedEdit: Counterfactual Diffusion-Based Image Editing on Brain MRI},
  author={Alaya, Malek Ben and Lang, Daniel M and Wiestler, Benedikt and Schnabel, Julia A and Bercea, Cosmin I},
  booktitle={International Workshop on Simulation and Synthesis in Medical Imaging},
  pages={167--176},
  year={2024},
  organization={Springer}
}

@inproceedings{xia2024mitigating,
  title={Mitigating attribute amplification in counterfactual image generation},
  author={Xia, Tian and Roschewitz, M{\'e}lanie and De Sousa Ribeiro, Fabio and Jones, Charles and Glocker, Ben},
  booktitle={MICCAI},
  year={2024},
}

@article{roschewitz2024robust,
  title={Robust image representations with counterfactual contrastive learning},
  author={Roschewitz, M{\'e}lanie and Ribeiro, Fabio De Sousa and Xia, Tian and Khara, Galvin and Glocker, Ben},
  journal={arXiv preprint arXiv:2409.10365},
  year={2024}
}

@inproceedings{ilse2021selecting,
  title={Selecting data augmentation for simulating interventions},
  author={Ilse, Maximilian and Tomczak, Jakub M and Forr{\'e}, Patrick},
  booktitle={ICML},
  pages={4555--4562},
  year={2021},
  organization={PMLR}
}

@inproceedings{kumar2023debiasing,
  title={Debiasing counterfactuals in the presence of spurious correlations},
  author={Kumar, Amar and Fathi, Nima and Mehta, Raghav and Nichyporuk, Brennan and Falet, Jean-Pierre R and Tsaftaris, Sotirios and Arbel, Tal},
  booktitle={FAIMI-MICCAI},
  pages={276--286},
  year={2023},
  organization={Springer}
}

@article{pan2024counterfactual,
  title={Counterfactual Image Editing},
  author={Pan, Yushu and Bareinboim, Elias},
  journal={arXiv preprint arXiv:2403.09683},
  year={2024}
}

@article{taylor2023,
	title = {Patient-specific modeling of blood flow in the coronary arteries},
	volume = {417},
	issn = {00457825},
	doi = {10.1016/j.cma.2023.116414},
	journaltitle = {Computer Methods in Applied Mechanics and Engineering},
	author = {Taylor, Charles A. and Petersen, Kersten and Xiao, Nan and Sinclair, Matthew and Bai, Ying and Lynch, Sabrina R. and {UpdePac}, Adam and Schaap, Michiel},
	year = {2023}
}

@inproceedings{kanitsar2002,
	location = {Boston, {MA}, {USA}},
	title = {{CPR} - curved planar reformation},
	isbn = {978-0-7803-7498-0},
	doi = {10.1109/VISUAL.2002.1183754},
	eventtitle = {{VIS}2002. {IEEE} Visualization 2002.},
	pages = {37--44},
	booktitle = {{IEEE} Visualization, 2002. {VIS} 2002.},
	publisher = {{IEEE}},
	author = {Kanitsar, A. and Fleischmann, D. and Wegenkittl, R. and Felkel, P. and Groller, E.},
	date = {2002},
}

@article{pegios2024diffusion,
  title={Diffusion-based iterative counterfactual explanations for fetal ultrasound image quality assessment},
  author={Pegios, Paraskevas and Lin, Manxi and Weng, Nina and Svendsen, Morten Bo S{\o}ndergaard and Bashir, Zahra and Bigdeli, Siavash and Christensen, Anders Nymark and Tolsgaard, Martin and Feragen, Aasa},
  journal={arXiv preprint arXiv:2403.08700},
  year={2024}
}

@inproceedings{roschewitz2024counterfactual,
  title={Counterfactual contrastive learning: robust representations via causal image synthesis},
  author={Roschewitz, M{\'e}lanie and de Sousa Ribeiro, Fabio and Xia, Tian and Khara, Galvin and Glocker, Ben},
  booktitle={DEMI-MICCAI},
  pages={22--32},
  year={2024},
  organization={Springer}
}

@inproceedings{ibrahim2024semi,
  title={Semi-Supervised Learning for Deep Causal Generative Models},
  author={Ibrahim, Yasin and Warr, Hermione and Kamnitsas, Konstantinos},
  booktitle={MICCAI},
  pages={294--303},
  year={2024},
  organization={Springer}
}

@inproceedings{puglisi2024enhancing,
  title={Enhancing spatiotemporal disease progression models via latent diffusion and prior knowledge},
  author={Puglisi, Lemuel and Alexander, Daniel C and Rav{\`\i}, Daniele},
  booktitle={MICCAI},
  pages={173--183},
  year={2024},
  organization={Springer}
}

@inproceedings{xia2025segmentor,
  title={Segmentor-Guided Counterfactual Fine-Tuning for Locally Coherent and Targeted Image Synthesis},
  author={Xia, Tian and Sinclair, Matthew and Schuh, Andreas and De Sousa Ribeiro, Fabio and Mehta, Raghav and Rasal, Rajat and Puyol-Ant{\'o}n, Esther and Gerber, Samuel and Petersen, Kersten and Schaap, Michiel and others},
  booktitle={MICCAI},
  pages={526--536},
  year={2025},
  organization={Springer}
}

@inproceedings{rasal2025diffusion,
  title={Diffusion Counterfactual Generation with Semantic Abduction},
  author={Rasal, Rajat R and Kori, Avinash and Ribeiro, Fabio De Sousa and Xia, Tian and Glocker, Ben},
  booktitle={ICML},
year={2025},
}

\end{document}